\documentclass{article}
\pdfoutput=1
\usepackage{ijcai23}

\usepackage{times}
\usepackage{soul}
\usepackage{url}
\usepackage[hidelinks]{hyperref}
\usepackage[utf8]{inputenc}
\usepackage[small]{caption}
\usepackage{graphicx}
\usepackage{amsmath}
\usepackage{amsthm}
\usepackage{amssymb}
\usepackage{epsfig}
\usepackage{booktabs}
\usepackage{algorithm}
\usepackage{algorithmic}
\usepackage[switch]{lineno}
\usepackage{empheq}
\usepackage{derivative}
\usepackage{booktabs, makecell, multirow}
\usepackage{comment}

\title{Instance Adaptive Prototypical Contrastive Embedding for Generalized Zero Shot Learning}

\author{
Riti Paul$^1$
\and
Sahil Vora$^1$\And
Baoxin Li$^1$
\affiliations
$^1$SCAI, Arizona State University\\
\emails
\{rpaul12, svora7, Baoxin.Li\}@asu.edu
}

\begin{document}

\maketitle

\begin{abstract}

    Generalized zero-shot learning(GZSL) aims to classify samples from seen and unseen labels, assuming unseen labels are not accessible during training. Recent advancements in GZSL have been expedited by incorporating contrastive-learning-based (instance-based) embedding in generative networks and leveraging the semantic relationship between data points. However, existing embedding architectures suffer from two limitations: (1) limited discriminability of synthetic features' embedding without considering fine-grained cluster structures; (2) inflexible optimization due to restricted scaling mechanisms on existing contrastive embedding networks, leading to overlapped representations in the embedding space. To enhance the quality of representations in the embedding space, as mentioned in (1),  we propose a margin-based prototypical contrastive learning embedding network that reaps the benefits of prototype-data (cluster quality enhancement) and implicit data-data (fine-grained representations) interaction while providing substantial cluster supervision to the embedding network and the generator. To tackle (2), we propose an instance adaptive contrastive loss that leads to generalized representations for unseen labels with increased inter-class margin. Through comprehensive experimental evaluation, we show that our method can outperform the current state-of-the-art on three benchmark datasets. Our approach also consistently achieves the best unseen performance in the GZSL setting\footnote{This paper won the Best Paper Award on IJCAI 2023 Workshop on \textit{Generalizing from Limited Resources in the Open World} (https://sites.google.com/view/glow-ijcai-23/schedule?authuser=0 )}.
    
\end{abstract}

\section{Introduction}

\label{sec:intro}

\begin{figure}[htpb]
    \centering
    \includegraphics[width=\linewidth]{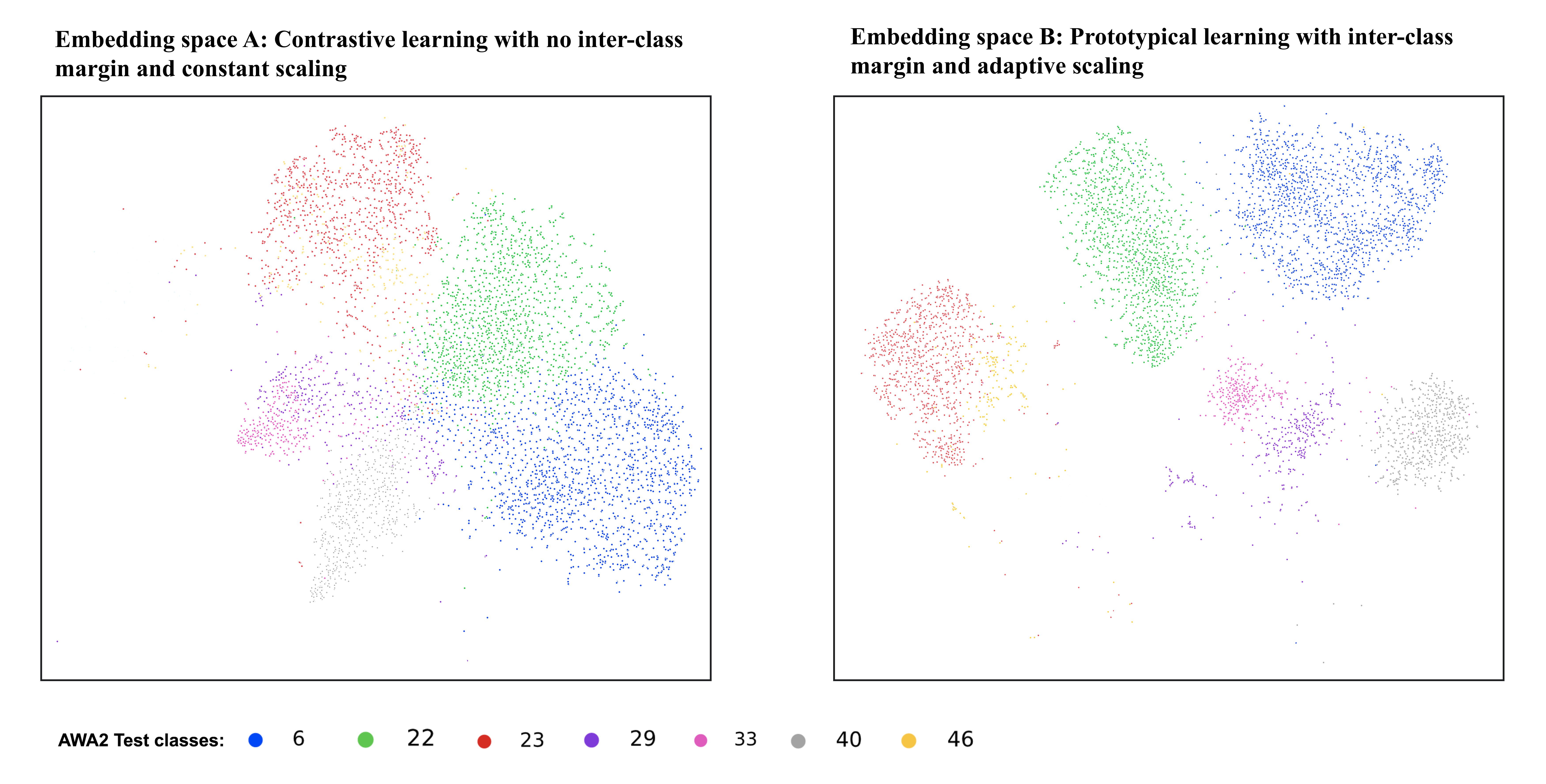}
    \caption{Comparison between the embedding representation of CE-GZSL(left) and our PCE-GZSL(right) for 7 unseen labels of AWA2. As can be observed, fine-grained representation learning may also lead to entangled and non-separable distributions. However, due to the nature of optimization in PCE-GZSL, it is able to learn fine-grained and well defined clusters with significant inter-class margin.}
    \label{fig:schematic}
\end{figure}

Zero-shot learning is the task of classifying objects into novel categories by leveraging the knowledge of the training classes and building a semantic bridge between the seen and the unseen classes in a common feature space. Seen and unseen classes are usually related in a high dimensional vector space (semantic space) where each class has a class-level attribute commonly known as the semantic vector, which can be visual or text-based. Conventional zero-shot learning implies that the test set only comprises examples from unseen classes. However, Generalized Zero-Shot Learning (GZSL) is a more challenging setting incorporating test samples from seen and unseen classes. Employing the connection between seen and unseen labels in the semantic space, early works \cite{zhang2016zero}, \cite{akata2015evaluation}, \cite{akata2015label}, \cite{akata2016multi}, \cite{al2016recovering}, \cite{demirel2017attributes2classname} in this domain were based on learning embedding-based networks capable of transforming data samples from the visual space to the semantic space. Although such methods performed impressively in the conventional setting, they fail to generalize in the challenging setting of GZSL due to the presence of bias towards the seen labels.

To tackle the generalization issue, generative methods \cite{narayan2020latent}, \cite{chen2021free} have experienced a rise in interest. The availability of synthesized unseen samples transforms the classification stage into a supervised learning problem, thereby eliminating the problem of lack of training samples. However, as accurately pointed out by \cite{han2021contrastive}, \cite{chen2021free}, the visual features extracted from ResNet101 \cite{he2016deep} are sub-optimal for the task of GZSL. Hence, there is a need to learn discriminative embedding features from the original visual space. Recent publications have utilized contrastive learning and disentangling of features \cite{ye2021disentangling}, \cite{chen2021semantics}, \cite{han2021contrastive} to learn improved and distinctive representations. 

Despite the improvement made by these methods, they suffer from certain limitations. Firstly, ambiguity in semantic information of different classes can often lead to overlapped synthesized clusters. Prior works \cite{han2021contrastive}, \cite{chen2021free} have used promising metric learning to learn fine-grained embedding spaces. However, they often lead to non-separable and entangled representations due to their fragility towards out-of-class distributions, as shown in Figure \ref{fig:schematic}. We argue that data-data interaction in contrastive frameworks needs to be improved to provide the embedding network with intense cluster supervision. To address this limitation, we propose a margin-based prototypical contrastive embedding network that reaps the benefits of prototype-based comparison (prototype-data) and implicit instance-based comparison (data-data). Since the prototypes are learnable components of our network and dynamic, it enhances the discriminative nature of the embeddings by discarding noise and learning shared features. We also introduce an inter-class similarity threshold as a margin in our proposed objective function. Unlike conventional instance-based contrastive losses, where each data point serves as an anchor, the proposed network treats each learnable prototype as an anchor. It is associated with all data in a batch which results in the implicit extraction of class-specific information between data points, resulting in fine-grained representations with greater inter-class separation between clusters (due to margin).

Secondly, in existing contrastive GZSL frameworks, the penalty on the similarity scores of every data-data pair is restricted to be equal ($\gamma_{ins}$). We argue that in an instance-level learning scenario, where the objective is to learn class-discriminating and high-quality features, scaling of similarity loss gradients should be instance-wise, depending on each pair-wise similarity score. This allows for a smoother optimization and avoids succumbing to sub-optimal features. 
To this end, we improve current contrastive GZSL frameworks with a re-scaling mechanism that allows significant penalty for hard pairs and modulates the penalty for easy and semi-hard pairs accordingly. Through experiments on benchmark datasets, we show that adding these simple modifications enhances the performance of existing contrastive frameworks.

In essence, this paper provides the following set of contributions: 
\begin{itemize}
    \item We propose a margin-based \textbf{P}rototypical \textbf{C}ontrastive \textbf{E}mbedding network for \textbf{GZSL} (\textbf{PCE-GZSL}), capable of retrieving fine-grained representations and well-defined cluster formations with increased inter-class margin. 
    \item We address an issue of inflexible optimization in current contrastive GZSL frameworks with an instance adaptive re-scaling mechanism that further enhances the performance.
    \item Extensive experimental results on four benchmarks 
    demonstrate the advantages of the proposed architecture over its baseline and current state-of-the-art methods, hence establishing new performance metrics for these benchmarks.
\end{itemize}

\begin{figure*}[htpb]
    \centering
    \includegraphics[width=0.8\linewidth]{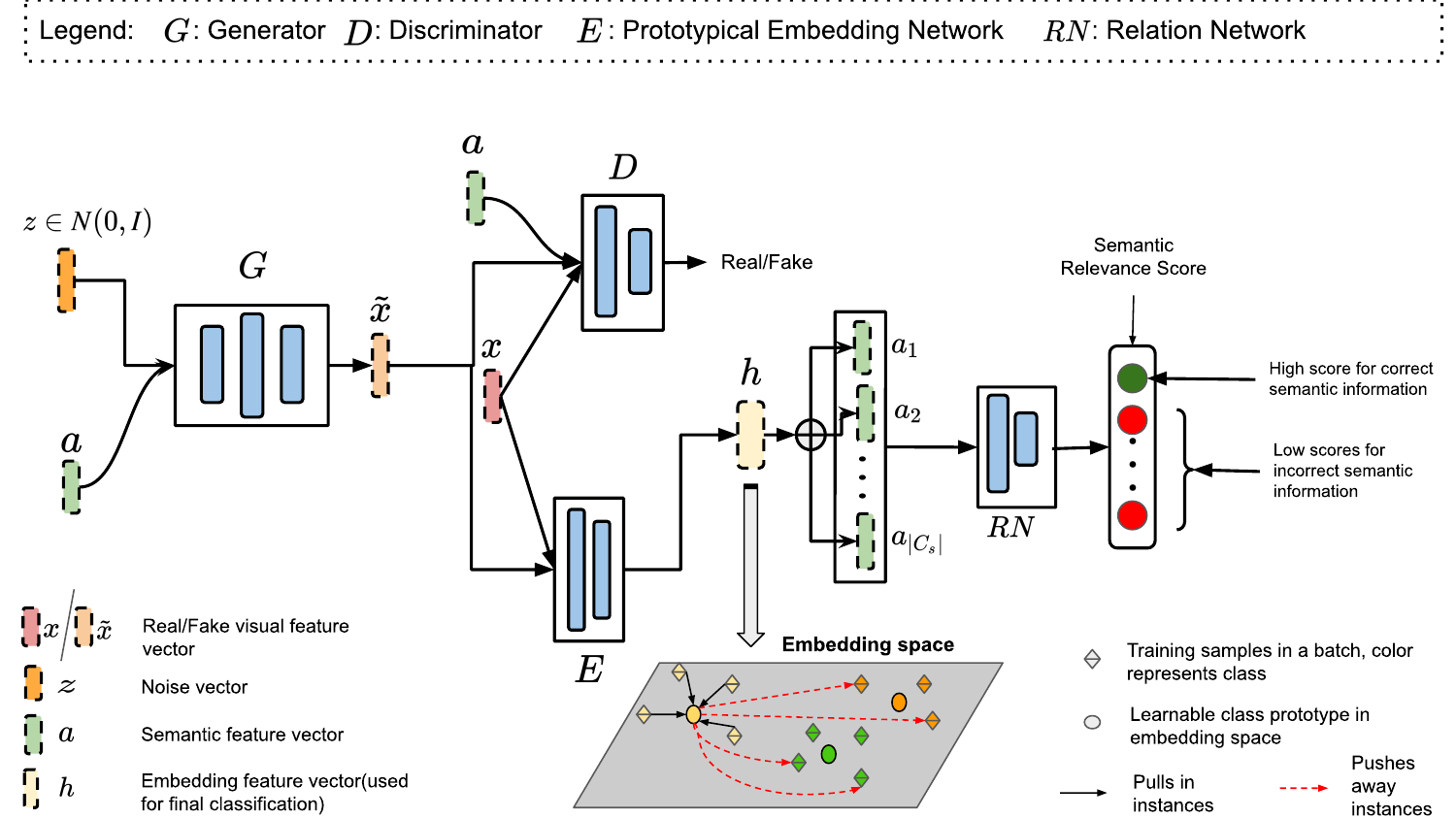}
    \caption{Illustration of our proposed method: \textbf{P}rototypical \textbf{C}ontrastive \textbf{E}mbedding (combined with WGAN architecture) for \textbf{G}eneralized \textbf{Z}ero Shot \textbf{L}earning (PCE-GZSL). Visual features(real/fake) are mapped to the embedding space using mapping network $E$. The embedding space is constrained by optimizing the instance adaptive prototypical contrastive loss in the projection space. We use dynamic prototypes with margin to 1)learn fine-grained embeddings using implicit instance interaction 2)provide stronger supervision to $E(.)$ about cluster information. We also use RelationNet, RN(.) to constrain the embedding space with semantic information.}
    \label{fig:proposed-method}
\end{figure*}

\section{Related Work}
\subsection{Zero Shot Learning}
In recent years, generative-based GZSL techniques have received much attention. They have shown promising results because of their greater capacity to overcome the bias and hubness issues frequently occurring in embedding-based GZSL techniques. The majority of the generative-based GZSL techniques were built on deep conditional generative models such as VAEs \cite{mishra2018generative}, \cite{schonfeld2019generalized}, GANs \cite{xian2018feature}, \cite{li2019leveraging}, \cite{10.1007/978-3-030-58577-8_5} or hybrid VAE/GAN \cite{xian2019f}, \cite{narayan2020latent} versions. Though existing methods have attained great success with GZSL, the original visual feature space lacks discriminative ability and is therefore sub-optimal for GZSL classification. 

\subsection{Contrastive Learning}
Contrastive learning has garnered much interest because of its promising performance in facilitating self-supervised computer vision tasks. Contrastive learning, also known as learning by comparing, has significantly enhanced self-supervised representation learning \cite{chen2020simple}, \cite{khosla2020supervised}. 
In \cite{han2021contrastive}, a GZSL approach based on contrastive
learning was proposed to map visual features to a more discriminative representation space by integrating the generative model
with the embedding model, yielding a hybrid GZSL framework. We consider CE-GZSL \cite{han2021contrastive} as the baseline for this work. 

\section{Proposed Methods}
\subsection{Problem Definition}
We are furnished with two sets of classes, seen and unseen, denoted by $S$ and $U$ respectively. The task of zero shot learning demands that $S \cap U = \phi$ to address the problem of long-tail recognition. Additionally, for every class $c_{i}$ in $Y = S \cup U$, a semantic embedding vector $a_{c_{i}}$ is provided, which retains the information on how the classes are related in the semantic embedding space. Let $D_{S}$ represent the set of labeled training data from the seen classes: $D_{S} = \{(x_i, c_i, a_{c_{i}}), x_i \in X_S, c_i \in S, a_{c_{i}} \in A, i = 1,2...N\}$, where $X_S$ is the set of $d$-dimensional image features from seen classes and $A$ is the set of semantic attribute vectors for all classes. The test set $D_{U} = \{x_{N+1}, x_{N+2}, \dots x_{N+M}\}$ is a set consisting of $M$ unlabeled images. The goal of zero shot learning is to construct a model $f_{ZSL}  :   x \rightarrow U$, that classifies samples from the unseen classes $U$ only. In the generalized zero shot learning setting, the objective is to learn a classifier $f_{GZSL} :   x \rightarrow U \cup S$ that can classify data from both the seen and unseen classes correctly. The latter setting is more difficult due to the presence of bias toward seen classes during classification phase which leads to confusion and may degrade the performance on either the seen or the unseen classes. 

\subsection{Generative Network - WGAN}
As the proposed architecture is a hybrid framework, in addition to a contrastive embedding network, we also use a feature generation network to address the missing sample problem of GZSL. The generative network(WGAN \cite{arjovsky2017wasserstein}) comprises of a conditional generator $G$ and a conditional discriminator $D$. The generator takes in a semantic descriptor and a randomly sampled noise vector from $N(0, I)$ as inputs to generate a visual feature $x'_{i}$. In the meanwhile, the discriminator attempts to classify the real visual feature($x_{i}$) from the synthesized feature($x'_{i}$), all conditioned on the semantic descriptor($a$). The WGAN loss optimizes both G and D,

\begin{equation}
    L_{WGAN} = \mathbb{E}[D(x, a)] - \mathbb{E}[D(x', a)] - \lambda \mathbb{E}[(|| \triangledown D(\hat{x}, a)||_{2} - 1)^{2}]
\end{equation}
\noindent where $\hat{x} = \tau x + (1 - \tau)x'$ where $\tau \in U(0,1)$ and $\lambda$ is the penalty coefficient.

\subsection{Prototypical Contrastive Embedding}
\subsubsection{Prototype-to-Instance level Embedding}
We use contrastive interaction between a learnable prototype and batch samples to learn \textit{fine-grained and well-clustered} representations. Each prototype is implemented as an embedding layer that updates itself based on its interaction with all batch samples. We compare each data point of a batch (positive/negative samples) against available prototypes (anchor).

The embedding network maps all visual features, real or synthesized, into an embedding space of dimensionality $D_{h}$ which can be formulated as $h_{i} = E(x_{i})$ where $h_{i} \in \mathbb{R}^{D_{h}}$. The embedding layer $E(.)$ is followed by a non-linear projection layer $H(.)$, which maps each $h_{i}$ into the projection space of dimensionality $D_{z}$. Therefore, $z_{i} = H(h_{i})$ or $z_{i} = H(E(x_{i}))$.
For any given dataset with $|S|$ training/seen classes(where $S$ denotes the set of seen classes), we define a set of prototypes $P$ such that $|P| = |S|$. For each prototype $z_{p}^{c} \in P$ where $c \in S$, the batch is divided into sets of $K$ negative samples(other classes) and $L$ positive samples(own class). We set up a contrastive problem which can be formulated as:

\begin{equation}
\begin{split}
L_{pr-ins} = \log(1 + \sum_{n=1}^{K}\exp(\gamma_{ins}(z_{p}^{c}z_{n}^{-} ))\\
\sum_{m=1}^{L}\exp(-\gamma_{ins}(z_{p}^{c}z_{m}^{+})))
\end{split}
\label{L-pr-ins}
\end{equation}

\noindent where $\gamma_{ins}$ is the scaling parameter, $z_{p}^{c} \in \mathbb{R}^{D_{z}}$  denotes the prototype for class $c \in S$ and $z^{+}_{m}, z^{-}_{n}$ represent the projection vectors of positive and negative samples respectively. The class prototype $z_{p}^{c}$ is updated dynamically during training, which helps capture class-specific features. 

In equation \ref{L-pr-ins}, there is no constraint on the similarity of prototype-data pairs. A plausible side-effect of this kind of optimization is entangled and non-separable representations in the embedding space, as shown in Figure \ref{fig:schematic}(a). We enforce between-class margin in our proposed loss function to learn well-separated, fine-grained representation. We introduce a new hyper-parameter $m$, known as the similarity margin. The margin $m$ controls the threshold value of the similarity score of any sample to the anchor(class prototype in this case). For any prototype $z^{c}_{p}$, $z^{c}_{p}z_{n}^{-} < m$ where $z_{n}^{-}$ is a negative sample and $z^{c}_{p}z_{m}^{+} > 1-m$ where $z_{m}^{+}$ is a positive sample. In other words, $m$ denotes the between-class margin($\delta_{n}$). Therefore, intra-class margin $\delta_{m} = 1-m$. Updating equation \ref{L-pr-ins} with margin $m$, we get the following:
\begin{equation}
\begin{split}
    L_{pr-ins}^{\delta} = \log(1 + \sum_{n=1}^{K}\exp(\gamma_{ins}(z^{c}_{p}z_{n}^{-} -\delta_{n}))\\
    \sum_{m=1}^{P}\exp(-\gamma_{ins}(z^{c}_{p}z_{m}^{+} - \delta_{m})))
\end{split}
\label{L-pr-ins-margin}
\end{equation}

In equation \ref{L-pr-ins-margin}, all similarity scores($z_{i}z_{n}^{-}, z_{i}z_{m}^{+}$) are restricted to be scaled by an equal quantity $\gamma_{ins}$. The implication of such restriction impacts the gradient and may adversely affect samples already close to their optimal similarity values. Let us take an example to illustrate this further - for a sample $z_{i}$, let us assume we have $z_{n}^{-}$ and $z_{m}^{+}$. The similarity scores for each of these samples are $z_{i}z_{n}^{-} = 0.9$ and $z_{i}z_{m}^{+} = 0.9$. This implies that the intra-class compactness is almost optimal and will only require smaller updates. However, the gradient for both $z_{n}^{-}$ and $z_{m}^{+}$ will be scaled by $\gamma_{ins}=80$(the value used in experiments). Although it is necessary for $z_{n}^{-}$(the higher penalty for sub-optimal values), scaling the gradient for $z_{m}^{+}$(already near-optimal) by $\gamma_{ins}$ seems unnecessary and may lead to sub-optimal decision boundaries.

To allow for a smoother optimization, depending on the current status of an instance, we propose an instance adaptive scaling mechanism where each similarity score is scaled by $\alpha^{i}_{m}=max(1+m - z_{i}z_{m}^{+}, 0)$(for positive samples) or $\alpha^{i}_{n}=max(z_{i}z_{n}^{-} - (-m), 0)$(for negative samples). The adaptive scaling helps to assign a greater penalty to hard samples and tone down the updates for an almost optimal pair. The $\alpha^{i}_{m}$ and $\alpha^{i}_{n}$ definitions have been adapted from \cite{sun2020circle} to reap the benefits of flexible optimization. Updating equation \ref{L-pr-ins-margin} with adaptive scaling mechanism, we arrive at the instance adaptive prototypical contrastive loss which is formulated as shown below

\begin{equation}
\begin{split}
    L_{pr-insAD}^{\delta} = \log(1 + \sum_{n=1}^{K}\exp(\gamma_{ins}\alpha^{c}_{n}(z_{p}^{c}z_{n}^{-} - \delta_{n})) \\
    \sum_{m=1}^{L}\exp(-\gamma_{ins}\alpha^{c}_{m}(z_{p}^{c}z_{m}^{+} - \delta_{m})))
\end{split}
\label{L-pr-insAD}
\end{equation}

\noindent where $\delta_{n}=m, \delta_{m}=1-m$ and $\alpha^{c}_{n}$, $\alpha^{c}_{m}$ are defined above. 

In equation \ref{L-pr-insAD}, it is easy to notice that for a prototype of class $c$, the optimizations will be modulated by a scale of $-\gamma_{ins}\alpha^{c}_{m}$ and $\gamma_{ins}\alpha^{c}_{n}$ for positive and negative samples, respectively. The gradient of equation \ref{L-pr-insAD} with respect to any sample not only contains information regarding its interaction with a prototype but also the relative hardness of all other samples in the batch. In this way, the samples in a batch interact with each other in a relative sense and assist in learning well-defined representations for instances and the class prototypes in the embedding space.

\subsubsection{Instance-to-Semantic level Embedding}
Learning fine-grained representations in an embedding space does not guarantee semantic relevance. Since features are generated from semantic attributes during the evaluation phase, constraining the embeddings with semantic information is crucial. This also applies to real visual features since the pre-trained feature space lacks discriminative features and is either ambiguous/coarse-grained.

For semantic-level distinction, the positive semantic attribute corresponds to the class $h_{i}$ belongs to, while the remaining ($S-1$) semantic attributes are treated as negative semantic embeddings. To accommodate dynamic interaction between $h_{i}$ and a set of available semantic attributes, we use RelationNet \cite{rusu2018meta} RN(.), which assigns a score for an embedding from the concatenation of $h_{i}$ and $a_{j} \forall j \in 1 \dots S$. The objective function for this problem is as follows:
\begin{equation}
    L_{sem} = - \log \frac{\exp(\gamma_{sem}RN(h_{i}, a^{+}))}{\sum_{j=1}^{S}\exp(\gamma_{sem}RN(h_{i}, a_{j}))}
    \label{L-sem}
\end{equation}

\begin{table}[tbp]
\centering
\resizebox{\linewidth}{!}{%
\begin{tabular}{|c|c|c|c|}
\hline
\textbf{Dataset} & \textbf{Total samples} & \textbf{Seen Labels} & \textbf{Unseen Labels} \\ \hline
AWA1 & 30,375 & 40 & 10 \\ \hline
AWA2 & 37,322 & 40  & 10 \\ \hline
SUN  & 14,340 & 645 & 72 \\ \hline
CUB  & 11,788 & 150 & 50 \\ \hline
\end{tabular}%
}
\caption{Dataset details}
\label{dataset}
\end{table}

\noindent where $\gamma_{sem}$ is a scaling parameter and $h_{i}$ denotes the embedding of sample $i$ of the batch. To enable substantial cluster supervision in the embedding feature space, we add a center semantic loss that helps the embedding network learn fine-grained clusters instead of overlapped class clusters. For this purpose, we formulate the following objective:
\begin{equation}
    L_{center-sem} = - \log \frac{\exp(\gamma_{sem}RN(\hat{h_{k}}, a^{+}))}{\sum_{j=1}^{S}\exp(\gamma_{sem}RN(\hat{h_{k}}, a_{j}))}
    \label{L-center-sem}
\end{equation}
\noindent where $\hat{h_{k}}$ denotes the prototype of embeddings of class $k \in S$. The primary motivation here is that along with samples from a class, the prototype of each class should also be semantically relevant. The total loss therefore is formulated as shown below:
\begin{equation}
\resizebox{\linewidth}{!}{$
L_{total} = L_{WGAN} + \lambda * L_{pr-insAD}^{\delta} + \beta * L_{sem} + \phi * L_{center-sem}
$}    
\label{L_total}
\end{equation}
\noindent where $\lambda, \beta, \phi$ are the scaling weights of the corresponding losses. The architecture of our proposed framework is depicted in Figure \ref{fig:proposed-method}. The embedding space is learnt from interactions between the class prototypes and the data. Semantic consistency in the embedding space is maintained with class-level supervision from the RN module. 

\begin{table*}[ht]
    \centering
    \resizebox{\textwidth}{!}{\begin{tabular}{||c | cccc | cccc | cccc | cccc||}
    \toprule
    \multirow{3}{*}[3pt]{Methods}
         &  \multicolumn{4}{c|}{\thead{AWA1}}
         & \multicolumn{4}{c|}{\thead{AWA2}} 
         & \multicolumn{4}{c|}{\thead{CUB}}
         & \multicolumn{4}{c||}{\thead{SUN}}\\
         \cmidrule(lr){2-5} \cmidrule(lr){6-9} \cmidrule(lr){10-13} \cmidrule(lr){14-17}
         & T & U & S & H & T & U & S & H & T & U & S & H & T & U & S & H \\
        \midrule
    DEVISE \cite{frome2013devise} & 54.2 & 13.4 & 68.7 & 22.4 & 59.7 & - & - & 27.8 & 52.0 & 23.8 & 53.0 & 32.8 & 56.5 & 16.9 & 27.4 & 20.9 \\
    DAP \cite{lampert2013attribute} & 44.1 & 0.0 & \textbf{88.7} & 0.0 & 46.1 & - & - & 0.0 
    & 40.0 & 1.7 & 67.9 & 3.3 & 39.9 & 4.2 & 25.1 & 7.2 \\
    SSE \cite{zhang2015zero} & 60.1 & 7.0 & 80.5 & 12.9 &  - & - & - & - & 43.9 & 8.5 & 46.9 & 14.4 & 51.5 & 2.1 & 36.4 & 4.0 \\
    ESZSL \cite{romera2015embarrassingly} & 58.2 & 6.6 &75.6 & 12.1 & 58.6 & - & - & 11.0 & 53.9 & 12.6 & 63.8 & 21.0 & 54.5 & 11.0 & 27.9 & 15.8 \\
    ALE \cite{akata2015label} & 59.9 & 16.8 & 76.1 & 27.5 & 62.5 & - & - & 23.9 & 54.9 & 23.7 & 62.8 & 34.4 & 58.1 & 21.8 & 33.1 & 26.3 \\
    LATEM \cite{xian2016latent} & 55.1 & 7.3 & 71.7 & 13.3 & - & - & - & - & 49.3 & 15.2 & 57.3 & 24.0 & 55.3 & 14.7 & 28.8 & 19.5 \\
    SYNC \cite{changpinyo2016synthesized} & 54.0 & 8.9 & 87.3 & 16.2 & 46.6 & - & - & 18.0 & 55.6 & 11.5 & 70.9 & 19.8 & 56.3 & 7.9 & 43.3 & 13.4 \\
    TCN \cite{jiang2019transferable} & 70.3 & 49.4 & 76.5 & 60.0 & 71.2 & 61.2 & 65.8 & 63.4 & 59.5 & 52.6 & 52.0 & 52.3 & 61.5 & 31.2 & 37.3 & 34.0 \\
    SE-GZSL \cite{verma2018generalized} & - & 56.3 & 67.8 & 61.5 & - & 58.3 & 68.1 & 62.8 & - & 41.5 & 53.3 & 46.7 & - & 30.5 & 40.9 & 34.9 \\
    f-CLSWGAN \cite{xian2018feature} & 68.2 & 57.9 & 61.4 & 59.6 & - & - & - & - & 57.3 & 43.7 & 57.7 & 49.7 & 60.8 & 42.6 & 36.6 & 39.4 \\
    CADA-VAE \cite{schonfeld2019generalized} & - & 57.3 & 72.8 & 64.1 & - & 55.8 & 75.0 & 63.9 & - & 51.6 & 53.5 & 52.4 & - & 47.2 & 35.7 & 40.6 \\
    f-VAEGAN-D2 \cite{xian2019f} & - & - & - & - & 71.1 & 57.6 & 70.6 & 63.5 & 61.0 & 48.4 & 60.1 & 53.6 & 64.7 & 45.1 & 38.0 & 41.3\\
    LisGAN \cite{li2019leveraging} & 70.6 & 52.6 & 76.3 & 62.3 & - & - & - & - & 58.8 & 46.5 & 57.9 & 51.6 & 61.7 & 42.9 & 37.8 & 40.2 \\
    TF-VAEGAN \cite{narayan2020latent} & - & - & - & - & 72.2 & 59.8 & 75.1 & 66.6 & 64.9 & 52.8 & 64.7 & 58.1 & 66.0 & 45.6 & \textbf{40.7} & 43.0 \\
    FREE \cite{chen2021free} & - & 62.9 & 69.4 & 66.0 & - & 60.4 & 75.4 & 67.1 & - & 55.7 & 59.9 & 57.7 & - & 47.4 & 37.2 & 41.7 \\
    HSVA \cite{chen2021hsva} & - & 59.3 & 76.6 & 66.8 & - & 56.7 & \textbf{79.8} & 66.3 & - & 52.7 & 58.3 & 55.3 & - & 48.6 & 39.0 & 43.3 \\
    CE-GZSL \cite{han2021contrastive} & 71.0 & 65.3 & 73.4 & 69.1 & 70.4 & 63.1 & 78.6 & 70.0 & 77.5 & 63.9 & 66.8 & 65.3 & 63.3 & 48.8 & 38.6 & 43.1 \\
    SCE-GZSL \cite{han2022semantic} & \textbf{71.5} & 65.1 & 75.1 & 69.7 & 69.4 & 64.3 & 77.5 & 70.3 & 78.6 & 66.5 & \textbf{68.6} & \textbf{67.6} & 62.5 & 45.9 & 39.7 & 42.7 \\ 
    DFTN \cite{jia2023dual} & 71.1 & 56.3 & 83.6 & 67.3 & 71.2 & 61.1 & 78.5 & 68.7 & 74.2 & 61.8 & 67.2 & 64.4 & - & - & - & - \\
    CvDSF \cite{zhai2023center} & 71.1 & 64.5 & 71.4 & 67.8 & - & 65.6 & 70.4 & 67.9 & 61.3 & 53.7 & 60.0 & 56.9 & 63.7 & 49.2 & 38.0 & 42.9 \\ \hline
    \textbf{Ours(PCE-GZSL)} & \textbf{71.5} & \textbf{67.2} & 73.8 & \textbf{70.3} & \textbf{73.9} & \textbf{67.0} & 74.8 & \textbf{70.6} & \textbf{79.2} & \textbf{66.9} & 65.8 & 66.4 & \textbf{66.8} & \textbf{51.3} & 37.9 & \textbf{43.6} \\
    \hline
    \end{tabular}}
    \caption{U and S are the average per-class Top-1 accuracies tested on unseen classes and seen classes, respectively, in GZSL. H is the harmonic mean of U and S. T represents the average per-class Top-1 accuracy for conventional zero shot learning setting. Best is in bold.}
    \label{tab:gzsl}
\end{table*}

\section{Results and Discussion}
\subsection{Datasets and Evaluation Metrics}
We compare our model with several state-of-the-art methods on 4 benchmark datasets: SUN \cite{patterson2012sun}, CUB \cite{wah2011caltech}, AWA1 \cite{lampert2009learning} and AWA2 \cite{xian2018zero}. We test our method on both coarse-grained (AWA1, AWA2) and fine-grained (CUB, SUN) datasets. Each category of AWA1/2 is annotated with an 85-dimensional semantic embedding. A 102-dimensional semantic embedding is used to annotate each category in the SUN dataset. For AWA1, AWA2 and SUN, we use the class embeddings provided by \cite{xian2018zero}. The semantic embeddings of CUB are 1024-dimensional feature vectors extracted by CNN-RNN \cite{basiri2021abcdm}.   For all the datasets, we use the ResNet-101 features provided by \cite{xian2018zero} as the image features. Remaining dataset related details have been summarised in Table \ref{dataset}. Finally, to evaluate the performance of the trained model, we adopt the widely used average per-class top-1 accuracy for conventional zero-shot setting. To measure the performance in GZSL setting, we use the harmonic mean $H$ of $U$ (average per-class accuracy on test images
from unseen classes) and $S$ (average per-class accuracy on
test images from seen classes). Once the networks are trained, in order to predict the label for test samples, we first generate samples for unseen class. Finally, we utilize the real seen samples and the synthetic unseen samples in the embedding space to train a softmax model as the final GZSL classifier. For ZSL, we only use synthetic samples for unseen labels to train the final classifier.

\subsection{Implementation details}

We choose Adam as our optimizer for all the networks, with hyperparameters $\beta_{1}=0.5, \beta_{2}=0.99$. The learning rate is $1\epsilon^{-4}$ for AWA1, AWA2, CUB and $5\epsilon^{-5}$ for SUN. AWA1 and AWA2 experiments used a batch size of 4096; CUB used 1024 and SUN 512 as their batch sizes. All networks are implemented as feed-forward MLP networks. The RelationNet RN(.) is a multi-layer perceptron (MLP) with a hidden layer followed by LeakyReLU activation. The relational network RN takes as input the concatenation of an embedding h and a semantic descriptor $a$ and outputs the relevance estimation between them. Our generator G and discriminator D both contain a 4096-unit hidden layer with LeakyReLU activation. For the embedding network $E(.)$, all datasets used 2048 as the embedding dimension($D_{h}$). However, for projection dimension($D_{z}$), AWA1, AWA2 performed better with 1024, CUB and SUN with 512. The hardware and
software information of our implementation is summarized as follows: (1) Software: Python 3.8, Pytorch 1.2, (2) GPU: NVIDIA Tesla V100 and (3) Operating System: Ubuntu 18.04. 

For all the datasets, we have used $\gamma_{ins}=80$, inter-class margin $m=0.4$ and $\gamma_{sem}=10$. For AWA1 and AWA2, $\lambda = \beta = \phi = 0.001$. For CUB and SUN, we achieved the best results with the following scaling weights : $\lambda = \beta = \phi = 0.01$. For our experiments we have generated 1800, 2400, 100 and 100 synthetic samples (for each unseen label) for AWA1, AWA2, CUB and SUN respectively.

\begin{figure*}[htpb]
    \centering
    \includegraphics[width=1.0\textwidth]{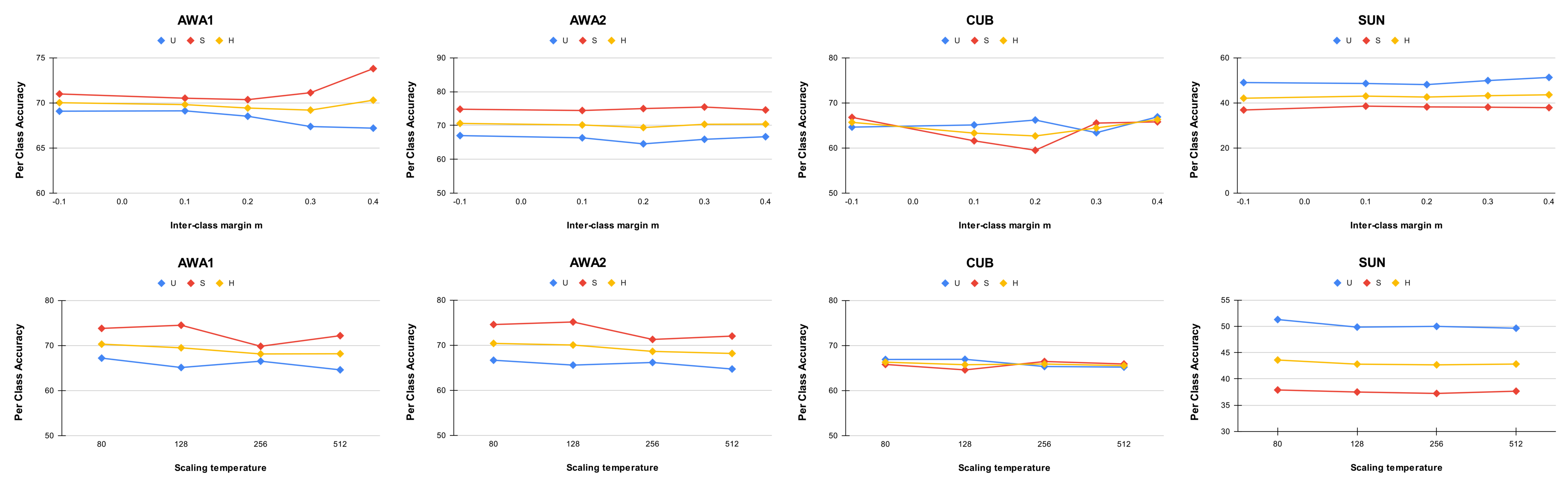}
    \caption{GZSL performance with respect to different values of inter-class margin, $m$ and scaling temperature $\gamma_{ins}$.}
    \label{fig:ablation-m}
\end{figure*}

\subsection{Comparison with state-of-the-art}
To demonstrate the effectiveness of our proposed approach (PCE-GZSL), we compare it against 20 other GZSL state-of-the-art methods. Table \ref{tab:gzsl} demonstrates the comparison results. We note that our proposed method achieves the highest H values on 3 of the datasets(AWA1, AWA2, SUN) and achieves the second-best performance for the CUB dataset in terms of H. Only contrastive algorithms are able to perform competitively with our proposed method, indicating that contrastive metric learning has certain advantages in the GZSL setting. For the CUB dataset, SCE-GZSL \cite{han2022semantic} performs better than our method, which could be contributed to the high number of latent projections(=number of features in semantic attributes) that the authors use in their method. The architecture of SCE-GZSL becomes highly non-scalable for datasets with a high number of attribute features(1024 attributes for CUB $\rightarrow$ 1024 latent projections!). Our method consistently achieves the best unseen label classification performance on all benchmark datasets, thereby reducing the bias towards seen classes and enhancing generalization towards unseen data in the embedding space. The unseen label classification observes an improvement of $1.9\%$, $2.7\%$, $0.4\%$ and $2.5\%$ in per class accuracy for AWA1, AWA2, CUB and SUN respectively. Our method also outperforms the top-1 per class unseen label accuracy for AWA2($1.7\%$), CUB($0.6\%$), SUN($0.8\%$) and is competitive on AWA1. 

\subsection{Ablation Study}
To provide detailed insight into instance adaptive PCE-GZSL, we conduct ablation studies of different components of the proposed network and evaluate their effects. Due to lack of space, we have included results for one dataset from coarse and fine-grained categories.

\noindent \textbf{Analysis of Model Components:} 

In this section, we investigate the contributions provided by the different levels of supervision proposed in our network. Table \ref{tab:ablation-comp} shows the performance comparison between different elements of our method, mainly (1) $L_{pr-insAD}^{\delta}$ (2) $L_{sem}$ (3) $L_{center-sem}$ and (4) $L_{sem} + L_{center-sem}$ when combined with the generative network loss $L_{WGAN}$. "$L_{WGAN} + L_{pr-insAD}^{\delta}$" denotes the prototype-instance level contrastive learning in PCE-GZSL. "$L_{WGAN} + L_{sem}$" and "$L_{WGAN} + L_{center-sem}$" represent the class-level supervision on each sample and cluster centers in the embedding space, respectively. Table \ref{tab:ablation-comp} shows that each element has a significant contribution except "$L_{WGAN} + L_{center-sem}$". This can be explained by the limited comparisons that are performed (only clusters centers in the embedding space) that constrain the capacity of the contrastive $RN(.)$ module. However, when both $L_{center-sem}$ and $L_{sem}$ are combined, there is a marginal improvement due to increased comparisons, making contrastive learning effective. Table \ref{tab:ablation-comp} also shows that the proposed prototypical contrastive learning outperforms the instance-based contrastive method for both $H$ and $U$. Furthermore, the proposed method significantly outperforms the baseline for unseen accuracy, $U$, in instance-level comparison (rows 1 and 2), thereby proving our claim that prototype interactive provides two-fold benefits (fine-grained representations and well-defined cluster supervision) and helps to learn better discriminative features.

\begin{table}[ht]
    \centering
    \resizebox{\linewidth}{!}{%
    \begin{tabular}{||c | ccc | ccc ||}
    \toprule
    \multirow{3}{*}[3pt]{Methods}
         &  \multicolumn{3}{c|}{\thead{AWA2}}
         & \multicolumn{3}{c||}{\thead{CUB}}\\
         \cmidrule(lr){2-4} \cmidrule(lr){5-7}
         & U & S & H & U & S & H \\
        \midrule
         \begin{tabular}{@{}c@{}c@{}}Baseline: CE-GZSL \\ $L_{WGAN} + L^{ins}_{ce}$ \end{tabular} & 60.14 & 74.94 & 66.96 & 58.8 & \textbf{66.5} & 62.4\\ \hline
         \begin{tabular}{@{}c@{}c@{}}Ours: PCE-GZSL \\ $L_{WGAN} + L_{pr-insAD}^{\delta}$ \end{tabular} & 62.50 & 76.36 & 68.73 & \textbf{66.24} & 65.03 & \textbf{65.63} \\ \hline
         \begin{tabular}{@{}c@{}c@{}}Ours: PCE-GZSL \\ $L_{WGAN} + L_{sem}$ \end{tabular} & 64.02 & \textbf{76.57} & 69.73 & 62.7 & 63.3 & 63.0\\ \hline
         \begin{tabular}{@{}c@{}c@{}}Ours: PCE-GZSL \\ $L_{WGAN} + L_{center-sem}$ \end{tabular} & 60.02 & 73.28 & 66.73 & 59.5 & 61.4 & 60.1\\ \hline
         \begin{tabular}{@{}c@{}c@{}}Ours: PCE-GZSL \\ $L_{WGAN} + L_{sem} + L_{center-sem}$ \end{tabular} & \textbf{64.92} & 76.30 & \textbf{70.15} & 61.92 & \textbf{66.33} & 64.04 \\ \hline
    \end{tabular}%
    }
    \caption{Model components analysis for AWA2 and CUB. Our proposed prototypical loss performs significantly better when compared to the instance-based baseline, CE-GZSL.}
    \label{tab:ablation-comp}
\end{table}

\begin{figure*}[htpb]
    \centering
    \includegraphics[width=1.0\textwidth]{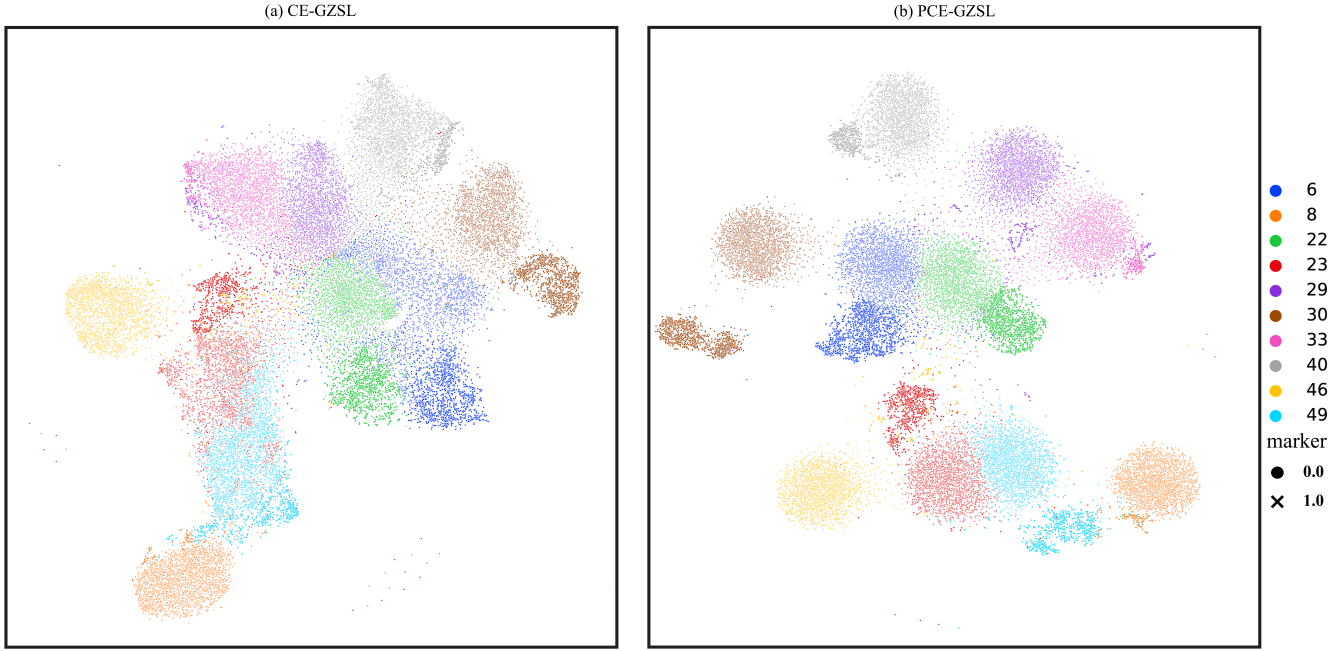}
    \caption{tSNE plot of classification features from embedding space of real and synthesized data for unseen labels for the dataset AWA2. The embedding space from CE-GZSL in (a) is unable to distinguish well between classes, due to entangled and non-separable distributions. In the embedding space, samples from the same class are closely clustered and different classes have significant margins between them in (b). The legend contains the color-coded labels of the unseen data and markers for distinguishing real (0) and fake (1) data.}
    \label{fig:cluster-tsne}
\end{figure*}
\noindent \textbf{Ablation of hyper-parameters:} To understand the impact of different hyper-parameters on the reported results, we conduct ablation for the following hyper-parameters: margin $m$, representing inter-class similarity and scaling temperature $\gamma_{ins}$. 

\begin{table}[ht]
    \centering
    \resizebox{\linewidth}{!}{%
    \begin{tabular}{||c | ccc | ccc ||}
    \toprule
    \multirow{3}{*}[3pt]{Methods}
         &  \multicolumn{3}{c|}{\thead{AWA2}}
         & \multicolumn{3}{c||}{\thead{CUB}}\\
         \cmidrule(lr){2-4} \cmidrule(lr){5-7}
         & U & S & H & U & S & H \\
        \midrule
        \begin{tabular}{@{}c@{}c@{}}Constant scaling, no margin \\ $L_{WGAN} + \lambda * L_{pr-ins}$ \\ $+ \beta * L_{sem} + \phi * L_{center-sem}$ \end{tabular} & 64.14 & 73.56 & 68.53 & 63.92 & 64.92 & 64.41 \\ \hline
        \begin{tabular}{@{}c@{}c@{}}Constant scaling, with margin \\ $L_{WGAN} + \lambda * L_{pr-ins}^{\delta}$ \\ $+ \beta * L_{sem} + \phi * L_{center-sem}$ \end{tabular} & 63.17 & \textbf{77.85} & 69.74 & 66.17 & 64.84 & 65.49 \\ \hline
        \begin{tabular}{@{}c@{}c@{}}Instance adaptive re-scaling, with margin \\ $L_{WGAN} + \lambda * L_{pr-insAD}^{\delta}$ \\ $+ \beta * L_{sem} + \phi * L_{center-sem}$ \end{tabular} & \textbf{67.0} & 74.8 & \textbf{70.6} & \textbf{66.9} & \textbf{65.8} & \textbf{66.4} \\ \hline

    \end{tabular}%
    }
    \caption{Contribution of the inter-class margin and re-scaling mechanism for datasets AWA2 and CUB}
    \label{tab:ablation_rescale}
\end{table}

\noindent \textbf{Inter-class margin, m:} The margin $m$ represents the inter-class margin, meaning $z^{c}_{p}z^{-}_{i} < m$ and $z^{c}_{p}z^{+}_{j} > (1-m)$ where $z^{c}_{p}$, $z^{-}_{i}$ and $z^{+}_{j}$ represent the prototype of class $c$ and its negative and positive samples respectively. For our study, we used the following values: $m = \{-0.1, 0.1, 0.2, 0.3, 0.4\}$. The top row of Figure \ref{fig:ablation-m} shows the GZSL performance(U, S, H) for all the datasets for the values as mentioned above of m(against $\gamma_{ins}=80, \gamma_{sem}=0.01$). Results show that the GZSL accuracy H is best when $m=0.4$ for all four datasets. However, for AWA1, we note that although H increases with $m=0.4$, the unseen accuracy decreases with rising $m$, indicating that it loses generalizability with higher m values. On the other hand, the fine-grained SUN dataset is robust to most of the margin values that have been experimented with.

\noindent \textbf{Scaling Temperature $\gamma_{ins}$:} In our proposed work, we employ the adaptive re-scaling mechanism implying that similarity scores are scaled by $\gamma_{ins}\alpha^{i}_{c}$, for instance, $z_i$ and prototype of class $c$. To understand the impact of the hyper-parameter $\gamma_{ins}$, we consider the following values for analysis: $\gamma_{ins} = \{80, 128, 256, 512\}$. Figure \ref{fig:ablation-m}(bottom row) shows the GZSL performance(U, S, H) for all the datasets for the values as mentioned above of $\gamma_{ins}$(against $m=0.4, \gamma_{sem}=0.01$). Due to the re-scaling factor $\alpha$, we can observe that accuracy $H$ is robust against the scaling hyper-parameter $\gamma_{ins}$.

\noindent \textbf{Effect of margin and re-scaling: } To understand the contributions of margin $m$ and instance adaptive re-scaling mechanism, we conduct experiments on AWA2 and CUB by substituting the prototypical loss term in equation \ref{L_total} with three different versions: (1) Constant scaling and no margin (Equation \ref{L-pr-ins}); (2) Constant scaling with margin $m=0.4$ (Equation \ref{L-pr-ins-margin}); (3) Instance adaptive scaling with margin $m=0.4$ (Equation \ref{L-pr-insAD}). Table \ref{tab:ablation_rescale} shows that the value of H is enhanced with both the margin and the re-scaling mechanism, implying that they lead to generalization to unseen data for any given dataset. 

\subsection{Visualization Results}
We visualize the classification features(from the embedding space) of unseen labels by first generating synthetic samples using the generator $G$, followed by embedding computation by the embedding module $E$. Embedding for real samples is computed by passing real test samples through embedding module $E$. Visualization of these embedding features in Figure \ref{fig:cluster-tsne} reveals the coarse-grained nature of unseen class clusters for our baseline CE-GZSL. However, the proposed method can form well-defined clusters and increase the distance between clusters of different classes, enhancing discriminability. 
\section{Conclusion}
In this work, we propose a GZSL framework based on a conditional generative model combined with a prototypical contrastive embedding network capable of learning semantically relevant fine-grained representations and well-defined cluster formations in the embedding space. We also point out that current contrastive GZSL frameworks suffer from inflexible optimization due to the constant scaling of similarity scores which we address in our work with the proposed instance adaptive loss. Experimental results exhibit the effectiveness of our proposed approach in mitigating some of the limitations in current GZSL frameworks. In our future work, we intend to learn a multi-modal embedding space with disentangled semantic information rather than constrain the embeddings with semantic relevancy and enhance the transferable knowledge in the discriminative embedding space by increasing the ability to comprehend unseen compositions of seen semantic attribute features. 

\bigskip

\noindent \textbf{Acknowledgement:} The work was supported in part by a grant from the National Science Foundation (Grant \#2126291). Any opinions expressed in this material are those of the authors and do not necessarily reflect the views of NSF.

\bibliographystyle{named}
\bibliography{ijcai23}

\end{document}